\documentclass[11pt]{article}

\usepackage{acl}

\usepackage{times}
\usepackage{latexsym}

\usepackage{algorithm}
\usepackage{algorithmic}
\usepackage{subcaption}   
\usepackage{graphicx}
\usepackage{tabularx}
\usepackage[utf8]{inputenc} % allow utf-8 input
\usepackage[T1]{fontenc}    % use 8-bit T1 fonts
\usepackage{url}            % simple URL typesetting
\usepackage{booktabs}       % professional-quality tables
\usepackage{amsfonts}       % blackboard math symbols
\usepackage{nicefrac}       % compact symbols for 1/2, etc.
\usepackage{microtype}      % microtypography
\usepackage{xcolor}         % colors
\usepackage{algorithmic}
\usepackage{algorithm}
\usepackage{multirow}     
\usepackage{booktabs}
\usepackage{enumitem}
\usepackage{amsmath}
\usepackage{makecell}

\usepackage[T1]{fontenc}
\usepackage[utf8]{inputenc}

\usepackage{microtype}

\usepackage{inconsolata}

\usepackage{graphicx}

\title{TLoRA: Task-aware Low Rank Adaptation of Large Language Models}
\author{
  Weicheng Lin\textsuperscript{*}, 
  Yi Zhang\textsuperscript{*}, 
  Jiawei Dang, 
  Liang-Jie Zhang\textsuperscript{$\dagger$} \\
  College of Computer Science and Software Engineering, Shenzhen University, China \\
  \texttt{2410105042@mails.szu.edu.cn, rambo.ai@szu.edu.cn, zhanglj@ieee.org}
}

\begin{document}
\maketitle
{
  \renewcommand{\thefootnote}{\fnsymbol{footnote}}
  \footnotetext[1]{These authors contributed equally to this work.}
  \footnotetext[2]{Corresponding author.}
}
\begin{abstract}
Low-Rank Adaptation (LoRA) has become a widely adopted parameter-efficient fine-tuning method for large language models, with its effectiveness largely influenced by the allocation of ranks and scaling factors, as well as initialization. Existing LoRA variants typically address only one of these factors, often at the cost of increased training complexity or reduced practical efficiency.
In this work, we present Task-aware Low-Rank Adaptation (TLoRA), a unified framework that jointly optimizes initialization and resource allocation at the outset of training. TLoRA introduces a data-driven initialization strategy that aligns the LoRA $A$ matrix with task-relevant subspaces by performing singular value decomposition on the product of pre-trained weights and input activation covariance. After this, the $A$ matrix is frozen, and only the $B$ matrix is trained. Furthermore, TLoRA employs a sensitivity-based importance metric to adaptively allocate ranks and scaling factors across layers under a fixed parameter budget.
We conduct extensive experiments that demonstrate TLoRA consistently performs excellently across various tasks, including natural language understanding, commonsense reasoning, math reasoning, code generation, and chat generation, while significantly reducing the number of trainable parameters. Our code is available at \url{https://github.com/Rambo-Yi/TLora/tree/main}
\end{abstract}

    \section{Introduction}

    Large Language Models (LLMs) exhibit remarkable capabilities across diverse Natural Language Processing (NLP) tasks \cite{achiam2023gpt, ouyang2022training, guo2025deepseek}. However, as model scales expand to billions or even hundreds of billions of parameters, full fine-tuning imposes prohibitive computational and memory overhead \cite{grattafiori2024llama, hoffmann2022training}. To address this, Parameter-Efficient Fine-Tuning (PEFT) has emerged as a solution that achieves performance comparable to full fine-tuning by updating only a minimal number of parameters, significantly reducing the requirements for domain-specific fine-tuning \cite{houlsby2019parameter, liu2021p, ding2023parameter}.
    
\begin{figure*}[t]
    \centering
    \includegraphics[width=0.6\linewidth]{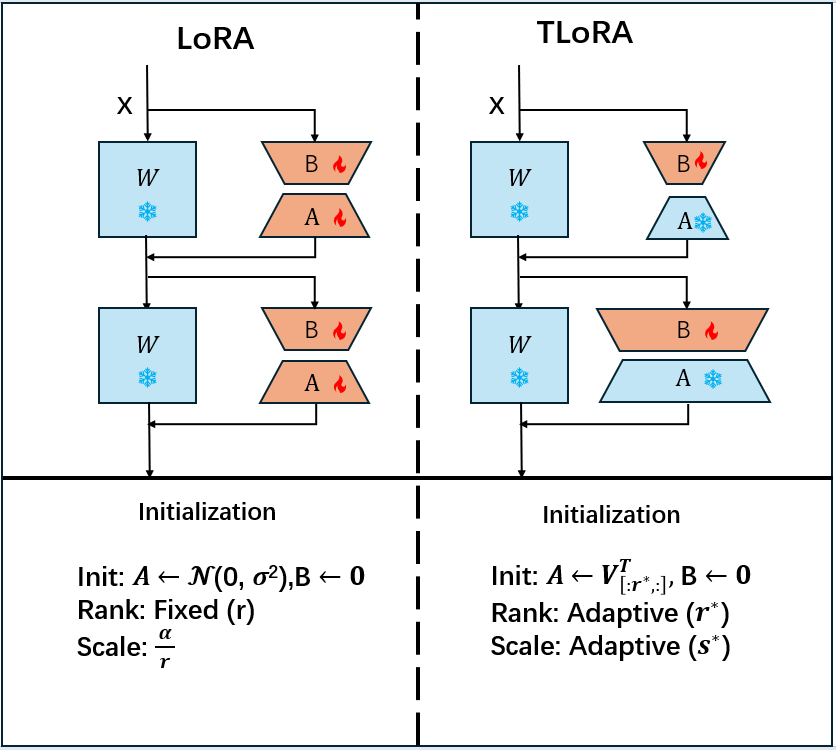} 
    \caption{{\bf (Left)} Standard LoRA employs random Gaussian initialization for matrix $A$ and initializes $B$ to zero, using a fixed rank $r$ across all modules. {\bf (Right)} TLoRA utilizes a task-aware initialization strategy where matrix $A$ is initialized using the top-$r^*$ singular vectors of the product of pre-trained weights and input covariance ($W_0C$) and is subsequently frozen (indicated by the snowflake). Furthermore, TLoRA adopts an adaptive resource allocation mechanism to dynamically determine the optimal rank $r^*$ and scaling factor $s^*$ for each module based on importance scores to ensure effective optimization.
    }
    \vspace{-0.45cm}
    \label{fig_init}
    \end{figure*}

    Among the various PEFT techniques, Low-Rank Adaptation (LoRA) \cite{hu2022lora} stands out as a notable approach. As depicted in Figure~\ref{fig_init} (Left), LoRA keeps the pre-trained weights $W$ frozen and approximates the weight update $\Delta W$ via two low-rank matrices, $A$ and $B$. In the standard implementation, $A$ is initialized with a Gaussian distribution, and $B$ is initialized to zero, ensuring that the initial model output remains identical to the pre-trained model. This design significantly reduces the number of trainable parameters and enables the product $BA$ to be merged into $W$ during inference, thereby avoiding extra inference latency. However, while LoRA performs well on simple tasks, recent studies indicate that it struggles to match the performance of full fine-tuning on complex tasks \cite{shuttleworth2024lora,biderman2024lora}.
    
    A significant factor influencing the effectiveness of LoRA is the initialization strategy. Fundamentally, matrix $A$ functions as a feature extractor that captures key low-rank information from high-dimensional inputs, while matrix $B$ serves as an output mapper that projects these features into the output space \cite{zhu2024asymmetry}. However, the random initialization of standard LoRA often results in a low-rank subspace defined by $A$ that is misaligned with task-relevant directions. This forces the model to consume substantial optimization steps at the beginning of training, primarily to rotate the projection subspace toward task-relevant directions, thereby hindering convergence efficiency. Motivated by this analysis, we hypothesize that if $A$ can be precisely aligned with task-relevant feature subspaces during the initialization phase, the feature extractor can remain frozen throughout subsequent training, requiring only the optimization of the mapper $B$ to achieve efficient adaptation.
    
    Another significant factor influencing the effectiveness of LoRA is the selection of rank and scaling factors. There are substantial differences in the contributions of different layers to downstream tasks. The uniform rank and scaling strategy of standard LoRA implicitly assumes that all layers contribute equally, which often leads to parameter wastage in non-critical layers while restricting the adaptation capacity of critical layers. Although most existing methods dynamically adjust the rank of each layer during the training process, this inevitably increases training overhead. More importantly, the $\alpha/r$ scaling strategy in standard LoRA actually limits the influence of critical layers when the rank varies. Therefore, we require a mechanism that avoids additional training costs and adaptively allocates resources based on module importance right at the initialization stage.
    
    In light of these challenges, we propose Task-aware Low-Rank Adaptation (TLoRA), as illustrated in Figure~\ref{fig_init} (Right). First, to address the initialization alignment issue, we introduce a data-driven initialization method. Specifically, we compute the input activation covariance matrix on a subset of training samples and perform Singular Value Decomposition (SVD) on the product of the pre-trained weight matrix and this activation covariance. The resulting right singular vectors are used to initialize and freeze matrix $A$; meanwhile, matrix $B$ is initialized to zero and remains the only trainable component. Second, to address the selection of rank and scaling factors, we utilize a sensitivity-based weight metric to assess the importance of each module and allocate ranks and scaling factors accordingly, ensuring that the resource budget is concentrated on the most critical components. Our contributions are summarized as follows:
    
    \begin{enumerate}[leftmargin=*, label=\arabic*.]
    \item  We derive the optimal closed-form solution $A^*$ under the constraint of a frozen projection matrix $A$. This result provides critical guidance for designing our initialization strategy.
    \item  We introduce an importance scoring mechanism that adaptively assigns different ranks and scaling factors to each module, enabling more efficient adaptation without increasing the parameter budget.
    \item We conduct comprehensive experiments across several challenging tasks, and the results demonstrate that TLoRA significantly outperforms LoRA and its variants on multiple benchmarks, even with an approximately 50\% reduction in parameter count.
    \end{enumerate}

%______________________________________________________________________
\section{Related Work}

\subsection{Parameter-Efficient Fine-Tuning (PEFT)}
PEFT methods aim to reduce computational and memory costs by fine-tuning only a small subset of model parameters while maintaining performance comparable to full fine-tuning \cite{ding2023parameter, xu2023parameter}. A variety of PEFT techniques have been proposed in recent years.
Adapter-based methods insert small trainable modules between the frozen layers of a pretrained model \cite{houlsby2019parameter, ruckle2020adapterdrop, pfeiffer2020adapterfusion}. While significantly reducing the number of trainable parameters, this approach typically incurs additional computational overhead, leading to increased inference latency.
Prompt-based methods add trainable vectors to the beginning of the input sequence to enable effective fine-tuning \cite{li2021prefix, lester2021power, liu2021p}. Although these methods demonstrate strong performance across a wide range of tasks, the extended input sequence length results in increased computational overhead during inference.
LoRA represents a particularly successful branch of PEFT methods \cite{hu2022lora}. It achieves effective adaptation by using low-rank decomposition matrices to approximate the weight updates, without introducing any additional inference latency after weight merging. This practical solution has inspired numerous extensions and improvements. Our work focuses on optimizing the initialization, rank, and scaling factor allocation to further unlock the potential of low-rank adaptation.

\subsection{LoRA Initialization}
LoRA initialization is crucial for convergence speed and final performance. Many studies show that Vanilla LoRA's reliance on random Gaussian initialization limits its ability to match Full Fine-Tuning on complex tasks. Current improvements to initialization primarily fall into two categories. 

The first category is weight-driven initialization. PiSSA performs Singular Value Decomposition (SVD) on pre-trained weights and initializes LoRA adapters using the principal singular vectors and values \cite{meng2024pissa}; in contrast, MiLoRA initializes adapters using the least significant singular vectors and values \cite{wang2025milora}; and OLoRA utilizes QR decomposition on pre-trained weights for initialization \cite{buyukakyuz2024olora}.

The second category is data-driven initialization. LoRA-GA aims to mitigate the initialization misalignment by aligning the adapter's initial gradients with those of Full Fine-Tuning \cite{wang2024lora}, and CorDA decomposes weights using context-oriented covariance matrices to preserve knowledge-dependent components \cite{yang2024corda}. While these methods enhance LoRA's performance across various tasks to some extent, they typically require modifying the pre-trained model weights ($W_{new}=W_{0}-A_0B_0$) to ensure that the model's output remains unaffected at the onset of training. This compromise undermines LoRA's advantage of efficient inference, as it requires either consuming significant time to reconstruct the initialization during inference or storing an additional set of weights for subsequent loading.

\subsection{Asymmetry and Adaptive Allocation}
Recent investigations into the internal mechanisms of LoRA have revealed a significant asymmetry between the two low-rank matrices: matrix $A$ is responsible for extracting features from the input, while $B$ projects these features onto the output \cite{zhu2024asymmetry}. This functional discrepancy has motivated a range of targeted optimization strategies. For example, LoRA+ \cite{hayou2024lora+} achieves enhanced performance by increasing the learning rate specifically for matrix $B$. LoRA-FA \cite{zhang2023lora} demonstrates that keeping $A$ frozen throughout the training process while training only matrix $B$ yields performance comparable to LoRA; this finding not only reduces memory overhead but also empirically suggests the feasibility of constructing a fixed, high-quality feature extractor $A$.

Furthermore, the selection of rank plays a decisive role in the performance of LoRA. It is widely acknowledged that increasing the rank dimension enhances the model's fitting capability. However, this improvement comes at the cost of linearly increasing GPU memory footprint and computational expense. Consequently, maximizing fine-tuning efficacy within a limited parameter budget has emerged as a focal point of current research. Recent works such as AdaLoRA \cite{zhang2023adalora} utilize SVD to iteratively prune singular values, achieving adaptive budget allocation across different layers and modules. Alternatively, DyLoRA \cite{valipour2023dylora} introduces a dynamic training objective that allows a single model to support multiple ranks simultaneously by training on a nested subset of low-rank matrices. However, a common limitation of this class of methods is their reliance on dynamic adjustments during training, which inevitably introduces additional computational overhead and significantly increases the complexity of the training pipeline.

% %______________________________________________________________________
\section{Method}
\label{sec:method}

\subsection{Background: Low-Rank Adaptation (LoRA)}

LoRA assumes that the weight updates required to adapt a pretrained model to downstream tasks lie in a low-dimensional subspace \cite{hu2022lora, aghajanyan2020intrinsic, li2018measuring}. Formally, for a given pretrained weight matrix $W_0 \in \mathbb{R}^{m \times n}$, LoRA approximates the update $\Delta W$ as follows:
\begin{equation}
  \label{eq:lora}
  W' = W_0 + \Delta W = W_0 + \frac{\alpha}{r} B A,
\end{equation}

where $W'\in\mathbb{R}^{m\times n}$, $A\in\mathbb{R}^{r\times n}$, $B\in\mathbb{R}^{m\times r}$, $r\ll\min(m,n)$, and $\frac{\alpha}{r}$ serves as a scaling factor for the update.

During fine-tuning, $W_0$ remains fixed, $A$ is initialized with a random Gaussian distribution, and $B$ is zero-initialized, ensuring that the model's outputs remain initially unchanged.

\begin{algorithm*}[t]
\caption{TLoRA Initialization}
\label{alg:TLoRA Initialization}
\begin{algorithmic}[1]
\STATE \textbf{Input:} Pre-trained model weights $\{W_i\}_{i=1}^L$, sample size N, LoRA rank \(r_{init}\), LoRA alpha \(\alpha\)
\STATE \textbf{Output:} Initialized LoRA modules $\{(A_i, B_i, \alpha_i, r_i)\}_{i=1}^L$.

\FOR{$i=1$ to \(L\)} 
    \STATE $S_i \leftarrow 0$ ,$C_i \leftarrow 0$
    
\ENDFOR

\FOR{$n =1$ to N}

\FOR{$i=1$ to \(L\)}
    \STATE $S_i \leftarrow S_i + \frac{1}{N}\ avg ( |W_i \cdot \nabla_{W_i} \mathcal{L}|) $
    \STATE $C_i \leftarrow C_i + \frac{1}{N}X_i^{\top}X_i$. 
\ENDFOR 
\ENDFOR
\STATE $R_{total} \leftarrow L \cdot r_{init}$, $\alpha_{total}\leftarrow L \cdot \alpha$
\FOR{$i=1$ to \(L\)}
    \STATE $r_i \leftarrow \left\lfloor R_{\mathrm{total}} \cdot \frac{\mathcal S(W_i)}{\sum_{j=1}^{L} \mathcal S(W_j)} \right\rceil$
    \STATE $\alpha_i\leftarrow r_i \cdot \frac{\alpha_{total}}{r_{init}} \cdot \frac{S(W_i)}{\sum_{j=1}^{L} S(W_j)} $ 
    \STATE $U, S, V^\top \leftarrow \text{SVD}(W_i C_i)$.
    \STATE $A_i \leftarrow V^\top_{[:r_i, : ]}$, $B_i\leftarrow 0$.
\ENDFOR

\RETURN $\{(A_i, B_i, \alpha_i, r_i)\}_{i=1}^L$
\end{algorithmic}
\end{algorithm*}
\subsection{Task-Aware Initialization}

To elucidate the critical role of matrix $A$ during the fine-tuning process, we first analyze the update dynamics of TLoRA subject to the constraint of a frozen $A$.

According to Eq.~\ref{eq:lora}, the gradient of the loss function $\mathcal{L}$ with respect to $B$ is $\frac{\partial \mathcal{L}}{\partial B} = \frac{\partial \mathcal{L}}{\partial W}A^T$. Therefore, during the training process, the update $\Delta W$ of the entire weight matrix can be expressed as:

\begin{equation}
  \label{eq:deltaW}
    \Delta W = \frac{\alpha}{r}\Delta BA
            =-\eta \frac{\alpha}{r}\frac{\partial \mathcal{L}}{\partial W}A^TA
\end{equation}

where $\eta$ represents the learning rate. Eq.~\ref{eq:deltaW} reveals a key mechanism: the weight update $\Delta W$ is strictly constrained within the subspace spanned by the row vectors of $A$.
This mechanism implies that if the row space of $A$ fails to effectively capture task-relevant feature directions, the model will be confined to a suboptimal subspace regardless of the optimization of $B$, thereby preventing performance from reaching the upper bound of full fine-tuning.

To determine the optimal initialization of $A$ under this constraint, we provide a detailed theoretical derivation in Appendix~\ref{theory_analysis}. We formulate the problem as minimizing the reconstruction error between the low-rank approximation and the ideal weight update. By solving this optimization problem, the closed-form solution is derived as:
\begin{equation}
A = V_{r}^{T}C^{-1/2} \in \mathbb{R}^{r \times n}
\end{equation}

where C is the input activation covariance matrix, and $V_{r}$ represents the top-$r$ right singular vectors derived from the Singular Value Decomposition (SVD) of $W_0 C^{1/2}$. However, directly applying this theoretical solution encounters severe numerical stability challenges. As detailed in Appendix~\ref{sec:appendix_stability}, the activation covariance matrix $C$ of LLMs typically exhibits an ill-conditioned spectrum characterized by a long tail of small eigenvalues and noise. In such scenarios, computing $C^{-1/2}$ necessitates inverting extremely small eigenvalues, which disproportionately amplifies task-irrelevant noise within the tail singular vectors, thereby causing extreme instability in the feature extractor during the initial training phase or even triggering gradient explosion.

To this end, TLoRA employs a robust approximation strategy. Instead of relying on unstable theoretical targets, we directly perform singular value decomposition (SVD) on the product of pre-trained weights and the covariance matrix ($W_0 C$), using the top-r right singular vectors to initialize the $A$ matrix.

Although $W_0 C$ and the theoretical target $W_0 C^{1/2}$ differ mathematically, our empirical analysis in Appendix~\ref{sec:appendix_alignment} indicates that they extract nearly identical feature subspaces. Specifically, we observe an average subspace similarity of 0.908 between the singular vectors of the two targets across all layers.

Moreover, theoretically, this approximation serves as a beneficial regularization effect. Since C weights feature directions based on variance ($\lambda$), while $C^{1/2}$ weights are based on standard deviation ($\sqrt{\lambda}$), using $W_0C$ imposes stronger penalties on tail noise directions. This effectively suppresses tail noise while preserving dominant task-specific directions validated by high subspace overlap.

Based on this analysis, the final initialization scheme is defined as:
\begin{equation}
    A = V_{r}^{T} \quad B = 0 
\end{equation}
where $V_{r}^{T}$ denotes the top-$r$ right singular vectors of the matrix $W_0C$. This approximation method effectively captures task-relevant directions while maintaining a high degree of numerical stability.

\subsection{Adaptive Rank and Scaling Assignment}

Our Task-Aware Initialization resolves the optimization of the projection subspace for an individual module. As derived in Appendix~\ref{theory_analysis}, for a given rank $r$, the maximum objective value achievable by the initialization matrix is determined by the sum of its top-$r$ task-specific singular values: $\sum_{i=1}^r \sigma_i(M_\Delta)$.

However, when extending our perspective to the entire model comprising $L$ modules, the global optimization objective becomes maximizing the total objective value across the system: $\sum_{n=1}^L \sum_{i=1}^{r_n} \sigma_i(M_\Delta^{(n)})$. The singular value spectra of $M_\Delta$ vary significantly across different modules, which fundamentally reflects their differing degrees of contribution to downstream tasks. Therefore, under a fixed total rank budget, a uniform rank assignment strategy is mathematically sub-optimal. To maximize the global optimization objective, the rank $r_n$ must be dynamically allocated, assigning larger rank capacities to modules with higher task importance. This reveals a tight theoretical coupling: initialization aligns the directions of the subspace, while rank allocation optimizes its capacity. Motivated by this unified global objective, we propose an adaptive allocation strategy that distributes the limited parameter budget to the more critical modules, thereby achieving more efficient model adaptation under the same budget.

To quantify each module's contribution, we employ a sensitivity-based importance metric \cite{liang2021super, zhang2022platon}. The underlying principle of this metric is that the most critical parameters are those with both a large magnitude and a significant impact on the loss. To define the importance score for a given module $i$, $\mathcal S(W_i)$, we first capture the influence of each parameter by calculating the product of its magnitude and gradient $|w \cdot \nabla_w \mathcal{L}|$, and then average these values across all parameters within the module:

\begin{equation}
    \label{eq:importance_score}
    \mathcal{S}(W_i) = \frac{1}{|W_i|} \sum_{w \in W_i} |w \cdot \nabla_w \mathcal{L}|
\end{equation}

where $w \in W_i$ denotes each parameter in the weight matrix $W_i$, $\nabla_w \mathcal{L}$ is the gradient of the loss with respect to $w$, and $|W_i|$ represents the number of parameters in $W_i$.
Given the total rank budget $R_{total}$ (sum of distributable ranks), we allocate a specific rank $r_i$ to each module $i$ proportional to its importance score:

\begin{equation}
\label{eq:rank_allocation}
    r_i = \left\lfloor R_{\mathrm{total}} \cdot \frac{\mathcal S(W_i)}{\sum_{j=1}^{L} \mathcal S(W_j)} \right\rceil
\end{equation}

Furthermore, to ensure that critical modules assigned a high rank can be effectively trained, it is necessary to adjust the scaling factor strategy. Standard LoRA employs a uniform scaling factor $s = \frac{\alpha}{r}$ across all modules. However, this uniform approach treats all modules indiscriminately, failing to account for disparities in their importance to the task. In our design, we concentrate the valuable rank budget on the critical modules most sensitive to the task. Adhering to the standard scaling of $\frac{\alpha}{r}$ would paradoxically cause these critical high-rank modules to receive smaller scaling factors, thereby inadvertently attenuating the update magnitude of the most critical modules.

Following the assignment of adaptive ranks based on importance scoring, we recalibrate the scaling factor to amplify the contributions of key modules. Specifically, instead of using a fixed $\alpha$, we dynamically calculate $\alpha_i$ for each module based on the importance score of the module:
\begin{equation}
\label{eq:alpha_allocation}
    \alpha_i = r_i \cdot \frac{\alpha_{total}}{r_{init}} \cdot \frac{\mathcal S(W_i)}{\sum_{j=1}^{L} \mathcal S(W_j)} 
\end{equation}

where $\alpha_{total}$ represents the total scaling budget across all layers, and $r_{init}$ denotes the initial rank (e.g., $r=8$). This formulation ensures that modules with higher importance scores $\mathcal S(W_i)$ are assigned both a larger rank $r_i$ and a correspondingly higher update magnitude $\alpha_i$, strictly adhering to our resource-concentration design objective. The detailed algorithm for TLoRA is in Algorithm~\ref{alg:TLoRA Initialization}.

% %______________________________________________________________________

\section{Experiments}
\label{sec:experiments}
We conduct a series of experiments to comprehensively evaluate TLoRA's performance across various scenarios. Initially, we evaluate its Natural Language Understanding (NLU) capabilities using the T5-base \cite{raffel2020exploring} model across a subset of GLUE \cite{wang2018glue}. Subsequently, to further evaluate TLoRA's performance in natural language generation (NLG), we perform extensive experiments on the LLaMA2-7B model \cite{touvron2023llama}, covering commonsense reasoning, math reasoning, code generation, and chat generation. Finally, we conduct an ablation study to validate TLoRA's effectiveness. We execute all experiments on a single NVIDIA A800 GPU. Hyperparameters are detailed in Appendix \ref{sec:hyperparameters}.

\subsection{Natural Language Understanding}
\noindent\textbf{Settings.} We evaluate TLoRA and baseline methods on the T5-base model. Our experimental evaluation covers multiple tasks from the GLUE benchmark, including MRPC \cite{dolan2005automatically}, CoLA \cite{warstadt2019neural}, RTE \cite{bentivogli2009rte5}, SST-2 \cite{socher2013sst2}, and QNLI \cite{rajpurkar2016squad}. We use accuracy as the evaluation metric across all tasks to ensure a consistent comparison.

\noindent\textbf{Results.} Table~\ref{tab:glue} shows the performance of TLoRA and several baselines on five representative tasks in the GLUE benchmark. TLoRA attains the highest scores on three of the five tasks (MRPC, COLA, RTE). While it maintains competitive performance on SST-2 and QNLI, TLoRA's average score (85.96\%) outperforms all baselines. These results substantiate TLoRA's effectiveness and versatility across diverse natural language understanding (NLU) tasks.
\begin{table}[ht]
    \centering
    \small
    \setlength{\tabcolsep}{1pt}
    \begin{tabular}{cccccccc}
   \toprule
    \multicolumn{1}{c}{\textbf{Method}}&{\makecell{Trainable \\ Parameters}} & \textbf{MRPC}  & \textbf{COLA}  & \textbf{RTE}   & \textbf{SST-2} & \textbf{QNLI}  & \textbf{Avg.} \\ \midrule
    FULL     &222.90M                           & 87.99          & 81.30          & 58.48          & 93.69          & \textbf{93.09}          & 82.91            \\
    LoRA      &12.97M                           & 85.53          & 76.03          & 63.17          &  94.49          & 93.00          & 83.44            \\
    AdaLoRA      &12.98M                           & 71.81          & 81.01          & 54.87          &  93.69          & 92.89          & 78.85            \\
    LoRA+    &12.97M           & 66.66          & 70.75          & 51.26         &  93.80       &  92.73        &  75.04          \\
    DoRA       &13.17M                         & 86.51          & 81.20          & 59.56          & 94.26          & 93.06 & 82.91            \\
    OLoRA       &12.97M                        & 87.99         & 75.83          & 66.78          & 90.13          & 90.48          & 82.24            \\
    PISSA       &12.97M                        & 88.72         & 81.20          & 69.67          & 94.03          & 92.60          & 85.24            \\
    LoRA-GA       &12.97M                        & 88.23         & 82.83          & 69.31          & \textbf{94.61}          & 93.00          & 85.59            \\
    CorDA       &12.97M                        & 88.48         & 81.78          & 71.11          & 93.92          & 92.75         & 85.60            \\
    TLoRA         &\textbf{5.44M}                       & \textbf{88.97} & \textbf{82.93} & \textbf{71.48} & 93.80 & 92.62          & \textbf{85.96}  \\\bottomrule
\end{tabular}
\caption{Results of fine-tuning T5-base using TLoRA and baseline method on a subset of GLUE. Bold numbers indicate the best performance achieved on this subtask. For TLoRA, the reported trainable parameters (5.44M) are the average across five tasks due to its adaptive rank mechanism.}
\label{tab:glue}
\end{table}

\begin{table*}[t]
\centering
\resizebox{1.0\textwidth}{!}
{
    \begin{tabular}{lcccccccccc}
    \toprule
    \multicolumn{1}{l}{\textbf{Method}} &{\makecell{Trainable \\ Parameters}}& \textbf{BoolQ} & \textbf{PIQA} & \textbf{SIQA}  & \textbf{HellaSwag} & \textbf{WinoGrande} & \textbf{ARC-e} & \textbf{ARC-c} & \textbf{OBQA}  & \textbf{Averge} \\ \midrule
    FULL       &6738M                         & 62.17          & 73.12          & 74.82          & 83.35     & 72.13               & 72.51 & 56.99          & 69.40          & 70.56           \\
    LoRA        &79.95M                        & 72.38          & 85.96          & 81.52          & 94.86     & 86.50               & \textbf{88.88} & 74.06          & 84.40          & 83.57           \\
    AdaLoRA   &79.96M                              & 71.34          & 83.51          & 81.62          & 93.74     & 84.68              & 87.58 & 73.37          & 83.80          & 82.46           \\
    LoRA+      &79.95M                           & 72.14         & 83.40          & 80.50          & 94.17     & 85.79              & 87.24 & 73.37          & 81.20          & 82.23           \\
    DoRA       &81.31M                         & 72.62          & 85.03          & 81.52          & 94.81              & 85.79               & 88.29          & 75.42 & 85.40          & 83.61           \\
    OLoRA     &79.95M                            & 71.34          & 85.90          & 80.91          & 94.15              & 85.39              & 87.07          & 74.57 & 84.00         & 82.92           \\
    PISSA    &79.95M                            & 71.89          & 85.03          & 80.96          & 93.91              & 85.47               & 86.78          & 73.72          & \textbf{86.40}          & 83.02           \\
    LoRA-GA    &79.95M                            & 69.93         & 84.33         & 81.26          & 94.08              & 85.47               & 87.07          & 72.18         & 84.40          & 82.34           \\
    CorDA    &79.95M                 & 67.33         & 79.86         & 79.22          & 91.60              & 83.10               & 82.82          & 68.00         & 81.40          & 79.17           \\
    TLoRA   &\textbf{41.68M}  & \textbf{72.87} & \textbf{86.28} & \textbf{82.59} & \textbf{95.21}              & \textbf{86.58}      & 88.59         & \textbf{76.36}          & 85.20 & \textbf{84.21}  \\ \bottomrule
    \end{tabular}
}
\caption{Evaluation results of commonsense reasoning of LLaMA2-7B on 8 tasks. Bold numbers indicate the best performance achieved on this subtask. }
    \label{tab:commonsense}
\end{table*}
\begin{table*}[t]
    \centering
    \resizebox{0.8\linewidth}{!}{
    \begin{tabular}{lcccccc}
    \toprule
    
    \multicolumn{1}{l}{\textbf{Method}} &{\makecell{Trainable \\ Parameters}} & \textbf{GSM8K} & \textbf{MATH} & \textbf{HumanEval} & \textbf{MBPP}  & \textbf{MT-Bench} \\ \midrule
    FULL      &6738M                           & 52.31          & 8.08          & 23.20              & 38.60          &   4.75               \\
    LoRA      &319.81M                          & 44.80          & 6.18          & 20.70              & 35.70          &    4.76               \\
    AdaLoRA    &319.84M                              & 43.20          & 5.74          & 20.70              & 36.00          &    4.50              \\
    LoRA+    &319.81M  & 48.67          & 6.92          & 22.60              & 35.40          &   4.69               \\
    DoRA     &321.17M                           & 45.10          & 5.96          & 20.70                   &  36.00              & 4.70                  \\
    OLoRA    &319.81M                            & 52.38          & 8.22          & 22.00                   &  38.90              & 4.99                 \\
    PISSA   &319.81M                            & 53.44          & 7.40          & 22.60              & 38.60          &  5.00                \\
    LoRA-GA   &319.81M                            & \textbf{58.15}          & 8.66          & 23.20              & 38.60          &  4.97                \\
    CorDA   &319.81M                         &54.43           & 8.70          & 21.30              & 39.40          &  5.09                \\
    TLoRA    &\textbf{171.71M}                           & 56.34 & \textbf{9.08} & \textbf{23.50}     & \textbf{40.20} &  \textbf{5.17}     \\ \bottomrule
    
    \end{tabular}
}
    \caption{Results of fine-tuning LLaMA2-7B using TLoRA and baseline method on math reasoning, code generation , and chat generation. Bold numbers indicate the best performance achieved on this subtask. For TLoRA, the reported trainable parameters (171.71M) are the average across three tasks due to its adaptive rank mechanism.}
    \label{tab:math_code_chat}
\end{table*}

\subsection{Natural Language Generation}
\noindent\textbf{Settings.} We conduct extensive experiments on the LLaMA2-7B model across four diverse task categories, each designed to evaluate a distinct capability of large language models: commonsense reasoning, math reasoning, code generation, and chat generation. These tasks serve as comprehensive benchmarks to evaluate the fine-tuning effectiveness of TLoRA in comparison with baseline methods.
\begin{itemize}[leftmargin=*, itemsep=0pt, parsep=0pt, topsep=3pt]
    \item \textbf{Commonsense Reasoning. }We fine-tune LLaMA2-7B on Commonsense170K \cite{hu2023llm}. We use accuracy as an indicator for commonsense reasoning tasks.
    \item \textbf{Math Reasoning. } We fine-tune LLaMA2-7B on a selected 100K subset of MetaMathQA \cite{yu2023metamath}. We evaluate on two benchmarks: GSM8K \cite{cobbe2021training} and MATH \cite{hendrycks2021measuring}. We measure the accuracy of the final answer.
    \item \textbf{Code Generation. }We fine-tune LLaMA2-7B on a 100K subset of Code-Feedback \cite{zheng2024opencodeinterpreter}. We evaluate on two benchmarks: HumanEval \cite{chen2021evaluating} and MBPP \cite{austin2021program}. Performance is measured using the pass@1 metric, representing the percentage of generated solutions that pass all test cases.
    \item \textbf{Chat Generation. }We fine-tune LLaMA2-7B on a 100K subset of WizardLM-Evol-Instruct \cite{xu2023wizardlm}. We evaluate using MT-Bench \cite{zheng2023judging}. Response quality is evaluated using GPT-4 as a judge.
\end{itemize}

\begin{figure}[t]
    \centering
    \includegraphics[width = 0.99\linewidth]{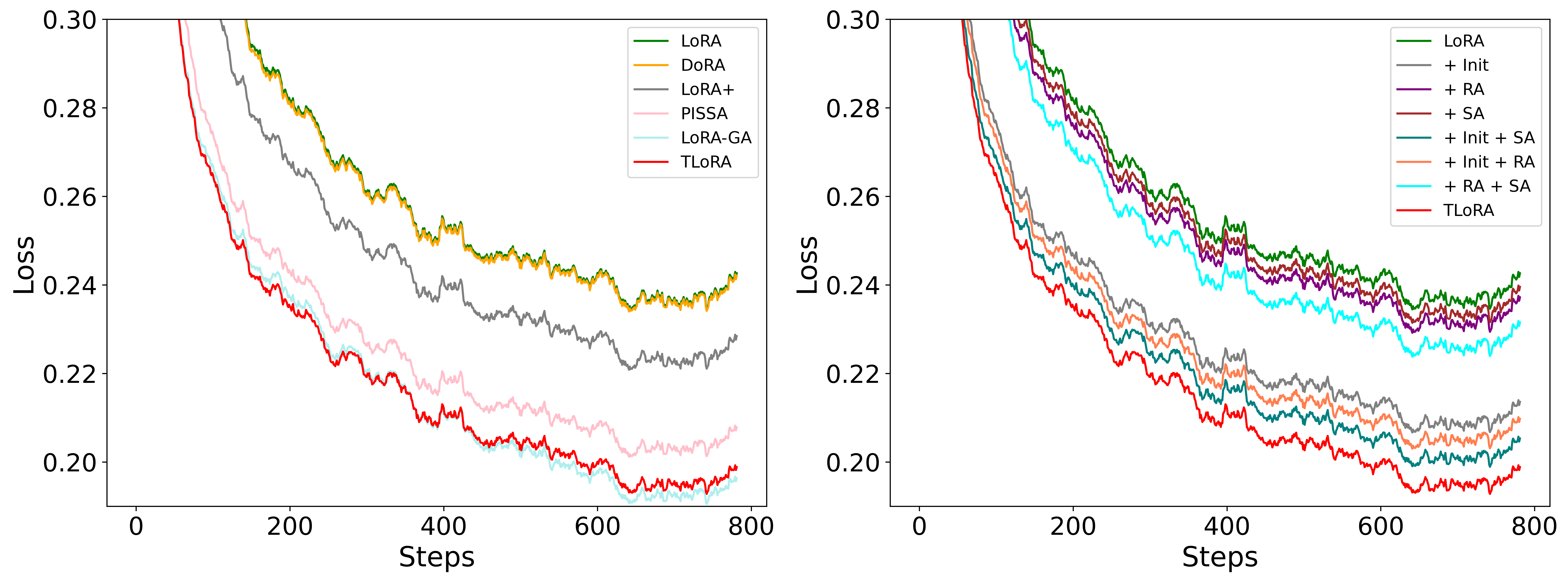}
        \caption{{\bf (Left)} Training loss curves of TLoRA and baseline with 128 ranks on the MetaMathQA dataset.  {\bf (Right)} Training loss curves from the ablation study with different settings on the MetaMathQA dataset.}
    \vspace{-0.45cm}
    \label{fig:loss}
\end{figure}

\noindent\textbf{Results.} The comprehensive experimental results are shown in Tables \ref{tab:commonsense} and \ref{tab:math_code_chat}. Notably, despite utilizing significantly fewer trainable parameters, TLoRA consistently achieves superior or highly competitive performance across various inference and generative benchmarks. On commonsense reasoning benchmarks, TLoRA demonstrates superior performance, achieving the best results on five out of eight subtasks and attaining the highest overall average accuracy (84.21\%). In the domain of math reasoning, TLoRA demonstrates robust capabilities. On the MATH math reasoning benchmark, TLoRA achieves an accuracy of 9.08\%, outperforming all baselines. For GSM8K, although it trails LoRA-GA marginally, TLoRA still achieves an excellent accuracy rate of 56.34\%, significantly surpassing Full Fine-Tuning (+4.03\%) and standard LoRA (+11.54\%). Similarly, in code generation tasks, which require both algorithmic reasoning and syntactic precision, TLoRA demonstrates clear superiority. It achieves the best performance on both HumanEval (23.50\%) and MBPP (40.20\%), outperforming all parameter-efficient baselines. Even in the highly complex domain of conversational AI, TLoRA surpasses all baseline methods and maintains optimal performance (5.17). Additionally, Figure~\ref{fig:loss} (Left) shows that our method maintains efficient convergence comparable to advanced initialization methods, while consistently outperforming standard LoRA. These results collectively highlight TLoRA as a highly efficient and effective fine-tuning strategy, delivering optimal performance across diverse tasks with reduced parameter overhead. Additionally, we demonstrate the effectiveness of TLORA across different model architectures and scales (e.g., LLaMA3 and Mistral) in Appendix \ref{app: more_model}.

\subsection{Ablation Study}
We conduct an ablation study to assess the impact of each component. The experimental setup is detailed as follows:
\begin{itemize}[leftmargin=*, itemsep=1pt, parsep=1pt, topsep=3pt]
\item \textbf{LoRA:} The standard LoRA.

\item \textbf{+ RA (Rank Adaptation):} The standard LoRA enhanced Rank Adaptation component.

\item \textbf{+ SA (Scale factor Adaptation):} The standard LoRA enhanced Scale factor Adaptation component.

\item \textbf{+ Init (Task-aware Initialization):} The baseline LoRA enhanced Task-aware Initialization component.

\item \textbf{TLoRA:} Our full proposed model, which integrates Init, RA, and SA.
\end{itemize}

\begin{table}[t]
    \centering
    \small
    \setlength{\tabcolsep}{4pt}
    \begin{tabular}{lccccc} 

    \toprule
    \textbf{Method} & \textbf{GSM8K}  & \textbf{MATH} & \textbf{HumanEval} & \textbf{MBPP} \\ \midrule
    \small
    LoRA           & 44.80         & 6.18   & 20.70  & 35.70             \\
    + RA          & 45.33          & 5.96   & 22.60 & 34.40            \\
    + SA          & 45.86          & 6.30   & 23.20 & 35.40           \\
    + Init        & 51.78          & 7.74   & 22.00 & 39.40           \\
    + Init + RA          & 54.05         & 7.68   & 22.60 & 38.60           \\
    + Init + SA        & 55.11          & 8.36   & 22.00 & 39.40           \\
    + RA + SA        & 47.68         & 6.50   & 22.60 & 36.50
    \\
    TLoRA           & \textbf{56.34} & \textbf{9.08}  & \textbf{23.50}  & \textbf{40.20}     
    \\ \bottomrule
    \end{tabular}
        \caption{Our ablation studies evaluate performance across various experimental settings. The results presented are derived from fine-tuning the LLaMA2-7B model using the Code Feedback 100k and MetaMathQA 100k data subsets.}
    \label{tab:ablation}
    %\vspace{-0.45cm}

\end{table}

The results of the ablation study (Table~\ref{tab:ablation}) perfectly corroborate our theoretical framework regarding the tight coupling between "direction alignment" and "capacity allocation." First, we observe that under standard random initialization, applying only rank adaptation (+RA) or scale factor adaptation (+SA) yields minimal gains and can even degrade performance. This is precisely because random initialization fails to align with the task-relevant feature basis; blindly allocating rank capacity or scaling updates within this "wrong" subspace fails to effectively capture core task-relevant features. In stark contrast, introducing only our task-aware initialization (+Init), leading to a substantial performance leap across all benchmarks. The full TLORA model successfully maximizes the global optimization objective through adaptive rank and scaling allocation, ultimately achieving the best overall performance. Furthermore, as shown in Figure~\ref{fig:loss} (Right), this integrated approach accelerates model convergence, underscoring its superior efficiency.

\begin{table*}[t]
    \centering
    
    \begin{tabular}{lcccccc}
    \toprule
    \textbf{Method} & \textbf{GSM8K} & \textbf{MATH} & \textbf{HumanEval} & \textbf{MBPP} &\textbf{MT-Bench}\\
    \midrule
    TLORA (Unfrozen-$A$) & \textbf{57.01} & 8.78 & 23.20 & \textbf{40.70} & 5.09\\
    TLORA (Frozen-$A$)   & 56.34 & \textbf{9.08} & \textbf{23.50} & 40.20 & \textbf{5.13}\\
    \bottomrule
    \end{tabular}
    
    \caption{Performance comparison between the standard TLORA (Frozen-A) and the fully trainable variant (Unfrozen-A) across math reasoning, code generation, and chat generation benchmarks. The results demonstrate that freezing $A$ maintains competitive performance while halving the trainable parameters per adapter.}
    \label{tab:frozen_vs_unfrozen}
\end{table*}

\begin{table*}[t]
    \centering
    \small
    \begin{tabular}{clccc}
    \toprule
    \textbf{Task} & \textbf{Method} & \textbf{1000 Steps} & \textbf{2000 Steps} & \textbf{3086 Steps (Final)} \\
    \midrule
    \multirow{2}{*}{GSM8K} & TLORA (Frozen $A$) & 59.05 & 62.47 & 64.36 \\
                           & TLORA (Unfrozen $A$) & 59.21 & 62.69 & 64.51 \\
    \midrule
    \multirow{2}{*}{MATH}  & TLORA (Frozen $A$) & 9.56  & 11.66 & 12.04 \\
                           & TLORA (Unfrozen $A$) & 9.26  & 11.92 & 13.08 \\
    \bottomrule
    \end{tabular}
    
    \caption{Performance of Frozen $A$ vs. Unfrozen $A$ on the MetaMathQA dataset across different training stages.}
    \label{tab:dynamic_evolution}
\end{table*}

\subsection{Expressiveness of the Frozen A Matrix}

A core design choice in TLORA is freezing the projection matrix $A$ post-initialization. Theoretically, freezing $A$ confines the weight update strictly to its initial row space, thereby restricting dynamic "subspace evolution." A natural concern is whether this limitation caps the model's expressiveness upper bound, particularly in complex tasks (e.g., mathematical reasoning or dialogue alignment) where the optimal low-rank subspace might evolve during training.

To rigorously address this concern, we provide evidence from both static final performance and dynamic training trajectories. First, as shown in Table~\ref{tab:frozen_vs_unfrozen},  the performance disparity is minimal across most benchmarks. Specifically, Frozen-$A$ not only matched the performance of the unfrozen variant on most tasks but even surpassed it on the MATH (9.08\% vs. 8.78\%), MBPP (23.50\% vs. 23.20\%), and MT-bench (5.13 vs. 5.09) benchmarks.

Second, to directly validate the necessity of "subspace evolution," we conducted a detailed tracking experiment on the complete MetaMathQA dataset. We compared Frozen-$A$ against the trainable Unfrozen-$A$ variant, evaluating their performance every 1,000 steps (Table~\ref{tab:dynamic_evolution}). The empirical data reveals that the performance gap between freezing and unfreezing $A$ remains extremely marginal across all training stages. This compellingly demonstrates that the initial subspace captured by TLORA via $W_0C$ is already near-optimal. Because this initialized subspace accurately aligns with the core feature directions required for complex reasoning from step zero, subsequent dynamic subspace evolution via gradient descent becomes largely redundant. Consequently, freezing $A$ is not a restrictive compromise, but rather a deliberate and highly effective design choice that guarantees maximum parameter efficiency and training stability without sacrificing task performance.

\subsection{Computational and Memory Analysis}
To validate TLoRA's efficiency, we fine-tune Llama-2-7B on the MetaMathQA dataset, recording actual training time and memory consumption. TLoRA requires a one-time precomputing phase for initialization and importance allocation. To minimize memory peaks during this phase, we compute the covariance matrix layer by layer and immediately offload it to CPU memory. As reported in Table~\ref{tab:time_cost}, the initialization process incurs merely 232 seconds, a negligible fraction of the approximately 5-hour total training duration. Furthermore, compared to LoRA, TLoRA freezes the projection matrix $A$, eliminating the need to store optimizer states for these parameters and significantly reducing GPU memory consumption during training.

\begin{table}[t]
\centering
\setlength{\tabcolsep}{4pt}
\begin{tabular}{c|ccc}
\hline
Method & \textbf{Stage} & \textbf{Time} & \textbf{Memory} \\ 
\hline
\multirow{2}{*}{LoRA} & Initialization & - & - \\
& Training & 4h48min15s & 63530MB \\
\midrule
\multirow{2}{*}{TLoRA} & Initialization & 232.47s & 17098 MB \\
& Training & 4h49min23s & 50448 MB \\
\hline
\end{tabular}
\caption{Time and GPU memory costs during initialization and training.}
\label{tab:time_cost}

\end{table}

% %______________________________________________________________________
\section{Conclusion}

This paper proposes TLoRA, a novel task-aware low-rank adaptation method that aligns adapters with task-relevant feature subspaces at the onset of training by performing SVD on the product of pre-trained weights and input activation covariance. Furthermore, to enhance parameter efficiency, we introduce a sensitivity-based importance scoring mechanism, enabling the adaptive allocation of ranks and scaling factors across different modules. Across extensive benchmarking, TLoRA consistently outperforms existing PEFT methods. Notably, since TLoRA operates solely during the initialization stage, it can be seamlessly integrated into existing LoRA pipelines as an efficient alternative initialization method, offering a simple yet effective enhancement to model adaptation.
% %______________________________________________________________________
\section*{Limitations}
\label{sec:limitation}

One limitation of this work stems from computational resource constraints, which restricted our evaluation primarily to the T5-Base and LLaMA2-7B models. Consequently, the scalability of TLoRA to larger-scale models (e.g., LLaMA2-70B) or Mixture-of-Experts (MoE) architectures remains to be empirically verified. 

A second limitation lies in the scope of our application; while our method is theoretically applicable to other architectures such as Vision Transformers (ViTs) and Vision-Language Models (VLMs), this study focuses exclusively on Natural Language Processing (NLP) tasks.

\section*{Acknowledgements}
This work was supported by the Guangdong Provincial Key Fields Special Project for Ordinary Universities (2025ZDZX1027).

\bibliography{custom}
\appendix

\section{Theoretical Derivation of Optimal Initialization}
\label{theory_analysis}

We theoretically analyze the effect of the selection of $A$ on fine-tuning performance. Inspired by the literature \cite{zhu2024asymmetry}, we formalize the problem as follows: given a projection matrix $A$, when $A$ is frozen and $B$ is optimized under the least-squares criterion, the expected loss corresponding to the optimal solution $B^*$ is denoted as $\mathcal{L}(A, B^*)$. Our objective is to identify the optimal $A^*$ that minimizes this loss. This loss function can be expressed as:
\begin{equation}
    \begin{aligned}
     \mathcal{L}(A,B^{\ast})  =\ &d_{out} \sigma^2 + \mathrm{Tr}[\Delta C \Delta^\top ] \\&- 
     \mathrm{Tr} [A C \Delta^\top  \Delta C A^\top  (A C A^\top )^{-1}]
    \end{aligned}
\end{equation}

Where $C = \text{Cov}[X]$ denotes the covariance matrix of the input data, $\Delta$ represents the weight update under ideal conditions, and $d_{out}\sigma^2$ represents the irreducible error. Since the first two terms, $d_{out}\sigma^2$ and $\text{Tr}[\Delta  C\Delta^\top]$, depend solely on the data distribution and the target task—being independent of the selection of $A$—our objective of minimizing the loss $\mathcal{L}$ is equivalent to maximizing the trace of the third term:
\begin{equation}
     J(A) = \mathrm{Tr} [A C \Delta^\top  \Delta C A^\top  (A C A^\top )^{-1}]
\end{equation}

To ensure numerical stability and ensure non-singularity, we employ Tikhonov regularization by introducing a small damping factor $\epsilon$ (e.g., $\epsilon=10^{-6}$): $C_{\text{reg}} = C + \epsilon I$. This modification guarantees that $C_{\text{reg}}$ is strictly positive definite. For notational simplicity in the following steps, we will denote the regularized matrix $C_{\text{reg}}$ simply as $C$.

Given that the covariance matrix $C$ is symmetric positive definite, there exists a unique symmetric positive definite square root $C^{1/2}$ such that $C = C^{1/2}C^{1/2}$. Let us define the transformed matrix as $\tilde{A} = AC^{1/2}$. Leveraging the cyclic property of the trace ($\mathrm{Tr}(ABC) = \mathrm{Tr}(BCA)$), we can rewrite the objective function in terms of $\tilde{A}$:
\begin{equation}
    \begin{aligned} 
    J(\tilde{A})  &= \mathrm{Tr} \left[ {\tilde{A}^\top (\tilde{A} \tilde{A}^\top)^{-1} \tilde{A}} \cdot {(\Delta C^{1/2})^\top (\Delta C^{1/2})} \right] 
    \end{aligned}
\end{equation}

Here, we decompose the optimization objective into the inner product of two matrices: $P_{\tilde{A}}$ and $M_{\Delta}$.
\begin{itemize}
\item $P_{\tilde{A}} = \tilde{A}^\top (\tilde{A} \tilde{A}^\top)^{-1} \tilde{A} $ represents the orthogonal projection operator onto the row space of $\tilde{A}$, constrained to rank $r$..
\item$ M_{\Delta} = (\Delta C^{1/2})^\top (\Delta C^{1/2})$ is a symmetric positive semi-definite matrix determined by the data and the ideal update quantity, containing the target directional information for the task.
\end{itemize}

Consequently, the optimization problem is transformed into finding a projection matrix $P_{\tilde{A}}$ of rank $r$ that maximizes $\mathrm{Tr}[P_{\tilde{A}}M_{\Delta}]$. 

To solve this extremum problem, we introduce the von Neumann trace inequality. For any two symmetric matrices $X$ and $Y$, the trace of their product satisfies the following upper bound:
\begin{equation}
    \label{Von Neumann's Trace Inequality}
    |\mathrm{Tr}(XY)| \le \sum_{i=1}^d \sigma_i(X) \sigma_i(Y)
\end{equation}

% 其中 $\sigma_i(·)$ 表示矩阵降序排列的特征值。 等号成立的充要条件是 $X$ 和 $Y$ 拥有相同的特征向量。
% 我们将此定理应用于当前优化目标,令 $X=P_{\tilde{A}}$,Y=$M_\Delta$。由于 $=P_{\tilde{A}}$ 是秩为 $r$ 的正交投影矩阵，其特征值$\sigma_i(P)$ 具有特定分布：对于 $1 \le i \le r$，$\sigma_i(P) = 1$，否则为 $0$。 因此，不等式简化为:

Where $\sigma_i(·)$ denotes the eigenvalues of the matrix sorted in descending order. The necessary and sufficient condition for the equality to hold is that $X$ and $Y$ share the same eigenvectors. We apply this theorem to the current optimization objective, setting $X=P_{\tilde{A}}$ and $Y=M_\Delta$. Since $P_{\tilde{A}}$ is an orthogonal projection matrix of rank $r$, its eigenvalues $\sigma_i(P)$ have a specific distribution: $\sigma_i(P) = 1$ for $1 \le i \le r$, and $0$ otherwise. Therefore, the inequality simplifies to:
\begin{equation}
    \mathrm{Tr}(P_{\tilde{A}}M_\Delta) \le  \sum_{i=1}^r \sigma_i(M_\Delta)
    \label{singular_M}
\end{equation}

This indicates that the maximum value of the objective function is determined by the sum of the top $r$ largest eigenvalues of $M_\Delta$. Based on the condition for equality in the inequality, the optimal projection matrix $P^*$ must share eigenvectors with $M_\Delta$. Specifically, $P^*$ must be the projection onto the eigensubspace corresponding to the top $r$ largest eigenvalues of $M_\Delta$.

Let the eigendecomposition of $M_{\Delta}$ be $M_{\Delta} = V \Sigma V^\top$, where $V_r \in \mathbb{R}^{n \times r}$ denotes the matrix composed of the top $r$ eigenvectors. Then the optimal projection matrix is uniquely determined as:
\begin{equation}
P^* = V_r V_r^\top
\end{equation}

Recalling the definition $P = \tilde{A}^\top (\tilde{A} \tilde{A}^\top)^{-1} \tilde{A}$, we note that $P$ depends solely on the row space of $\tilde{A}$ and not on the specific basis used to represent it. Consequently, any matrix $\tilde{A} = QV_r^T$ (where $Q \in \mathbb{R}^{r \times r}$ is invertible) will yield the same optimal projection matrix $P^* = V_rV_r^T$, resulting in the same optimal value for the objective function. Among all equivalent solutions, we select the canonical form where $Q=I_r$:
\begin{equation}
     \tilde{A}^* = V_r^T
\end{equation}

This derivation reveals a fundamental property: the theoretical optimal initialization is solely dependent on the principal directions (eigenvectors) of the target matrix $M_{\Delta}$. Consequently, the row space of the optimal adapter $\tilde{A}^*$ is uniquely spanned by the top $r$ right singular vectors of the target matrix $M_{\Delta}$.

In practical fine-tuning scenarios, the true update quantity $\Delta$ is completely unknown before training, which makes the direct construction of the target matrix $M_{\Delta}$ impractical. However, the theoretical derivation above establishes a crucial principle: the optimal projection matrix $P^*$ is solely determined by the principal eigendirections of the target matrix $M_{\Delta}$, independent of the specific magnitude of its eigenvalues. This implies that we do not need to reproduce the specific numerical values of $\Delta$; we only need to find a computable proxy matrix $M_{\text{proxy}}$ that exhibits high alignment with $M_{\Delta}$ in terms of its "subspace directions."

To construct this proxy matrix, we draw upon a key insight from the LoRA \cite{hu2022lora}: the weight update $\Delta$ resulting from fine-tuning exhibits a strong intrinsic correlation with the pre-trained weights $W_0$. Specifically, $\Delta$ does not randomly explore entirely new subspaces but instead tends to amplify certain directions that are already present in $W_0$ but were not fully emphasized. Building on this, we hypothesize that task-specific directions are essentially the pre-trained features ($W_0$) that are strongly activated by the current data ($C$). In other words, $W_0$ provides the potential feature basis, while $C$ acts as the "Selector" for filtering these directions. Consequently, we define the computable proxy matrix as: $M_{\text{proxy}} = (W_0 C^{1/2})^\top (W_0 C^{1/2})$. This formula assumes that the principal subspace of $M_{\text{proxy}}$ can serve as an accurate structural substitute for $M_{\Delta}$.

To verify the feasibility of this hypothesis, we conducted an empirical analysis using the Llama2-7B model on the MetaMath dataset. Specifically, we measured the subspace overlap between the proxy matrix $M_{\text{proxy}}$ and the ground-truth update matrix $M_{\Delta}$ obtained via full fine-tuning. We adopted the subspace similarity metric proposed in LoRA to measure this alignment:
\begin{equation}
    \label{eq:subspace_similarity}
    \phi(M_{\text{proxy}}, M_{\Delta}) = \frac{\| U_{\text{proxy}}^\top U_{\Delta} \|_F^2}{r} \in [0, 1]
\end{equation}
where $U_{\text{proxy}}$ and $U_{\Delta}$ denote the top-$r$ principal singular vectors of $M_{\text{proxy}}$ and $M_{\Delta}$, respectively. As illustrated in the figure, it depicts the subspace similarity between $M_{\Delta}$ and $M_{\text{proxy}}$ computed from the Q projection matrix of each layer in Llama2-7B. There is a significant subspace consistency between $M_{\text{proxy}}$ and $M_{\Delta}$, with the average similarity reaching 0.71. This evidence strongly supports our hypothesis that $M_{\text{proxy}}$ serves as an effective structural substitute for $M_{\Delta}$.
\begin{figure}[h]
    \centering
    \includegraphics[width=0.95\linewidth]{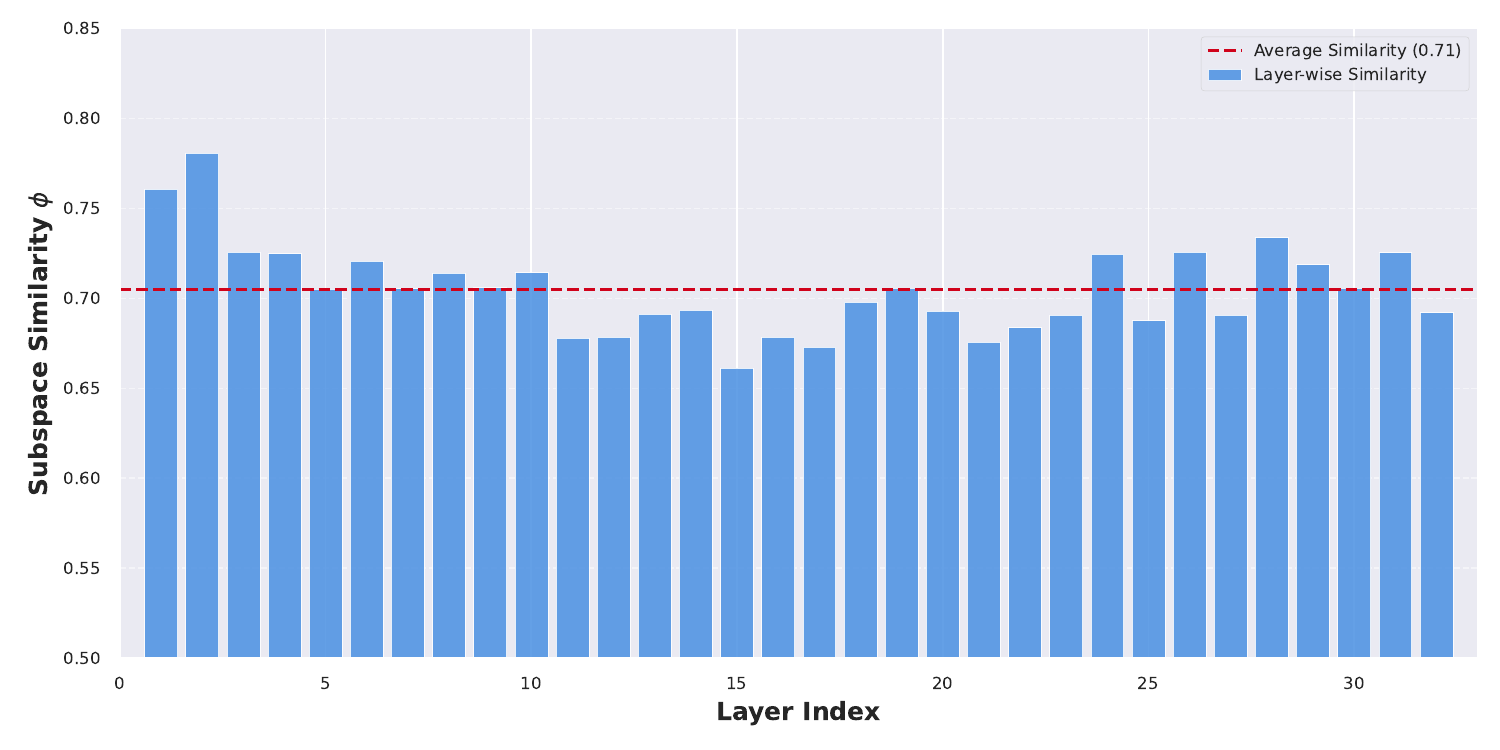}
    \caption{Subspace Similarity ($\phi$).}
    \label{fig:spectrum_analysis_theory}
\end{figure}

Finally, we map the optimization result back to the original parameter space via an inverse transformation to recover the LoRA projection matrix $A$. Recalling the transformation definition $\tilde{A} = A C^{1/2}$, by right-multiplying both sides of the equation by $C^{-1/2}$, we derive the theoretical optimal solution for $A$ as:
\begin{equation}
A = V_{r}^{T}C^{-1/2} \in \mathbb{R}^{r \times n}
\end{equation}

Where $V_r$ is the matrix composed of the top $r$ right singular vectors of the matrix $W_0 C^{1/2}$. This closed-form solution explicitly integrates both the pre-trained feature structure and data-driven activation statistics into the initialization of $A$.

\section{Empirical Analysis of Initialization Approximation}
\subsection{Numerical Instability of the Theoretical Solution}
\label{sec:appendix_stability}

Under the constraint of a frozen matrix $A$, we derived the theoretically optimal initialization solution $A = V_{r}^T C^{-1/2}$, where $V_{r}$ is obtained from the SVD of $WC^{1/2}$. However, in practice, TLoRA adopts a robust approximation strategy $A = V_{r}^T$, where $V_{r}$ is derived from the SVD of $WC$. In this chapter, we will provide a detailed analysis of the instability inherent in the theoretical solution.

\begin{figure}[h]
    \centering
    \includegraphics[width=0.95\linewidth]{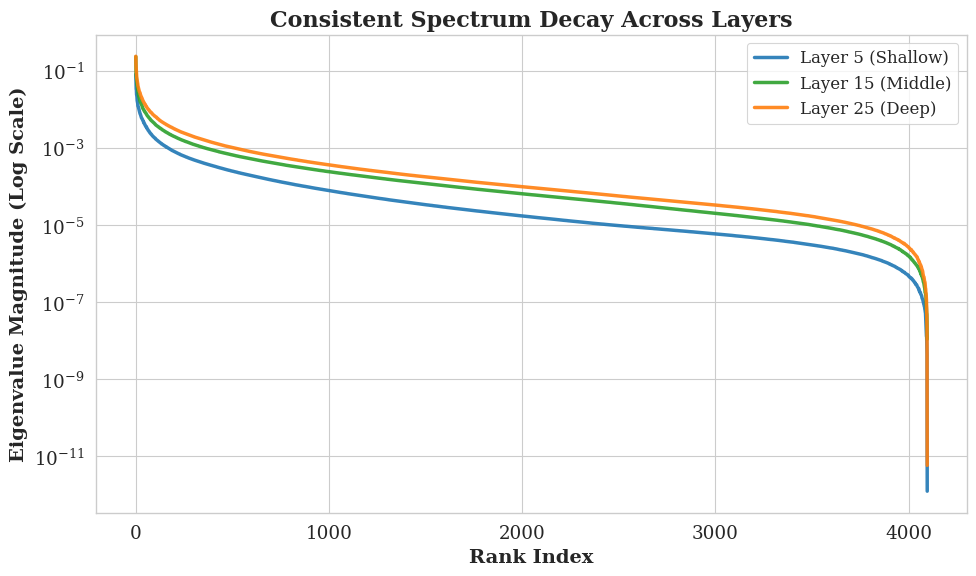}
    \caption{Eigenvalue spectrum consistency across varying layers.}
    \label{fig:spectrum_analysis}
\end{figure}

To investigate the root cause of the instability, we visualized the eigenvalues of the input activation covariance matrices for the $q$ matrices in layers 5, 15, and 25 of Llama-2-7B. As illustrated in Figure \ref{fig:spectrum_analysis}, the spectra across various layers exhibit a consistent long-tail distribution. The vast majority of these are extremely small eigenvalues. Inverting these tail values would systemically amplify noise, justifying our use of the robust approximation $A = V_r^T$.

To directly observe the instability of the theoretical solution, we compare the training performance of the theoretical solution with that of the approximate solution.

\begin{figure}[h]
    \centering
    \includegraphics[width=1.0\linewidth]{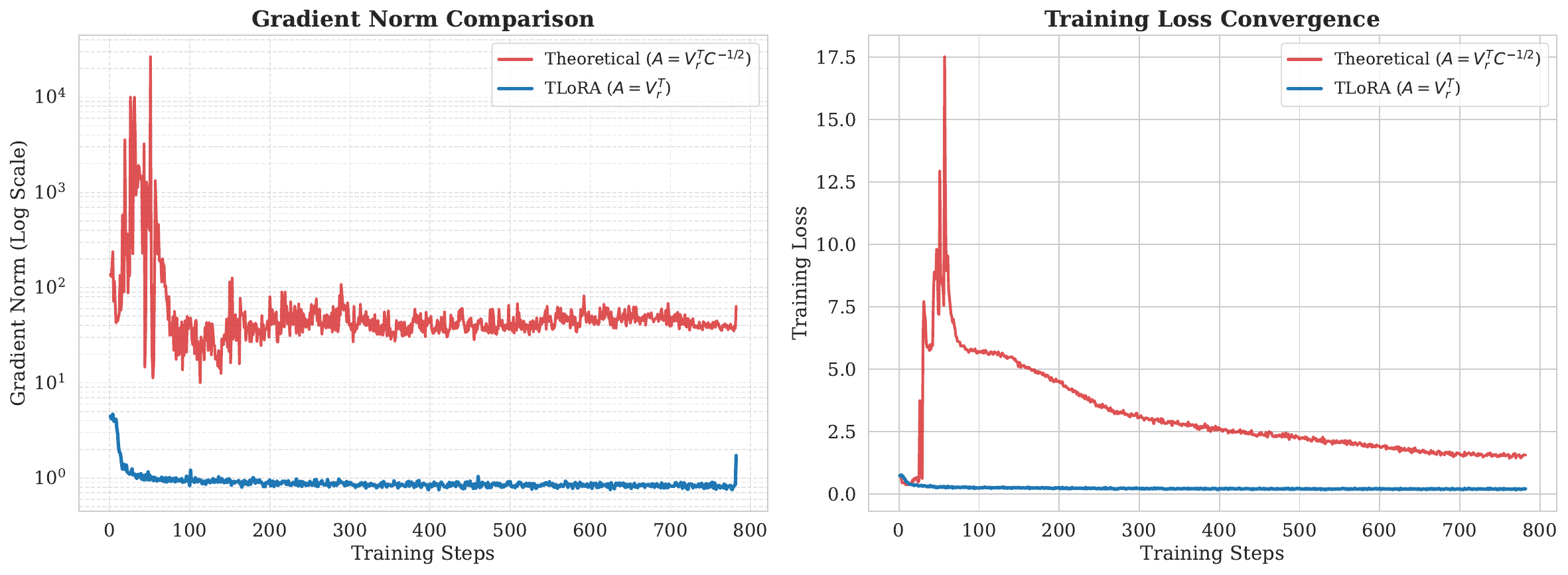}
    \caption{
    \textbf{(Left)} The gradient norm curves. 
    \textbf{(Right)} The training loss curves.}
    \label{fig:stability_comparison}
\end{figure}

Figure \ref{fig:stability_comparison} presents the gradient norm and training loss curves during the initial phase. The results provide compelling evidence: firstly, the theoretical solution exhibits a propensity for gradient explosion, with gradient norms exceeding those of TLoRA by several orders of magnitude. Secondly, this instability hinders effective model learning, resulting in a failure to converge. In contrast, TLoRA maintains stable and lower gradient norms, achieving rapid convergence.

\subsection{Subspace Alignment Analysis}
\label{sec:appendix_alignment}
Although theoretical derivations suggest extracting a sub-space from $W_0 C^{1/2}$, TLoRA actually performs SVD on $W_0 C$. We adopt this approximation to circumvent the computational cost and potential numerical issues associated with root extraction, while leveraging covariance's superior noise suppression properties compared to its square root. To validate that replacing $W_0 C^{1/2}$ with $W_0 C$ does not compromise the quality of the learned subspace, we conducted a quantitative analysis comparing structural alignment between the two objectives.

Specifically, we computed the subspace similarity $\phi$ between the top-$r$ right singular vectors derived from $W_0 C$ (denoted as $U_{Approx}$) and those derived from $W_0 C^{1/2}$ (denoted as $U_{Theory}$) across all layers of the LLaMA2-7B model. We utilized the projection metric defined as:
\begin{equation}
    \phi(U_{Approx}, U_{Theory}) = \frac{1}{r} \| U_{Approx}^\top U_{Theory} \|_F^2
\end{equation}

\begin{figure}[h]
    \centering
    \includegraphics[width=0.95\linewidth]{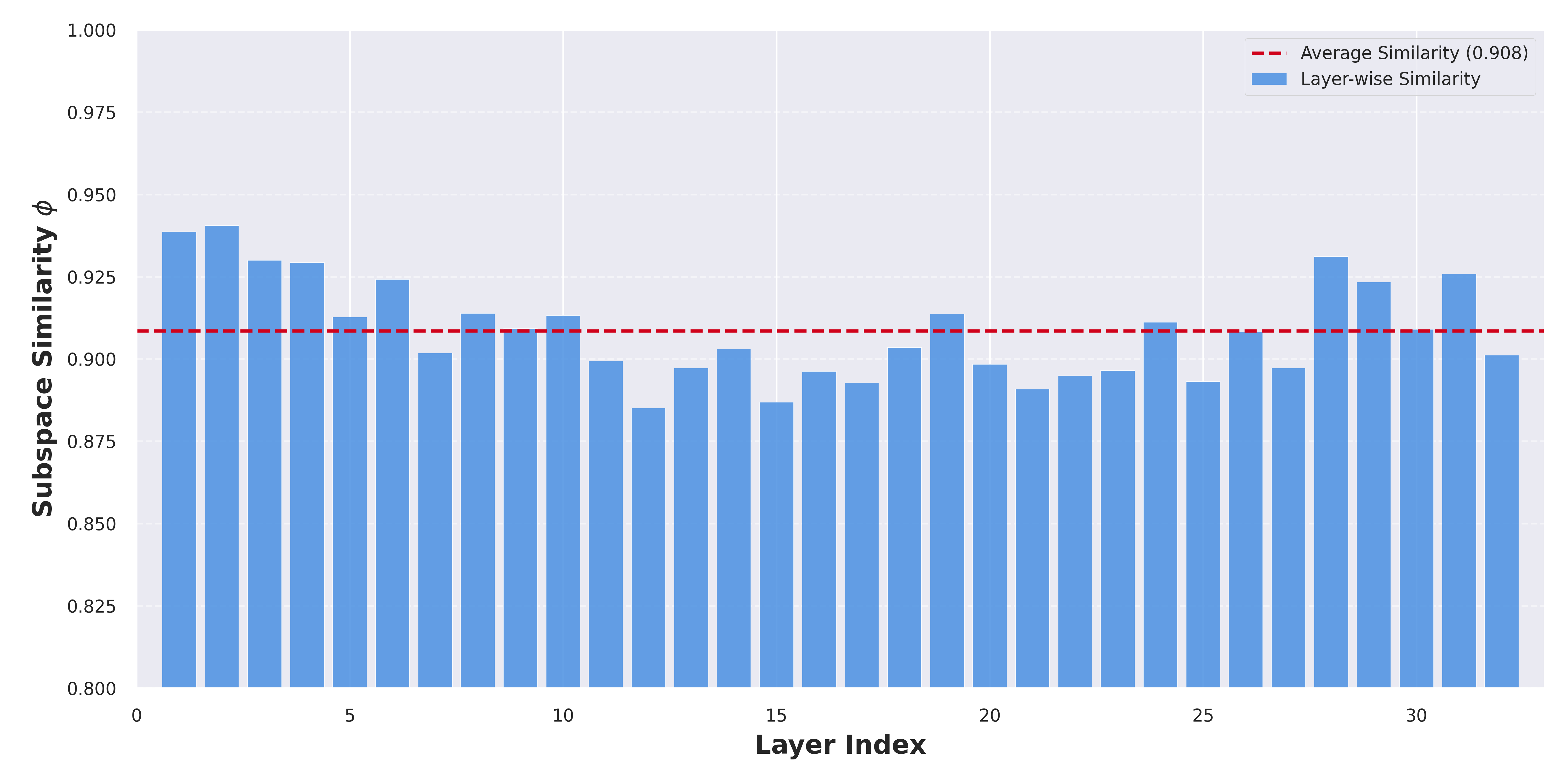}
    \caption{Subspace similarity analysis between the theoretical initialization target ($W_0 C^{1/2}$) and our robust approximation ($W_0 C$).}
    \label{fig:Subspace Alignment Analysis}
\end{figure}

As illustrated in Figure \ref{fig:Subspace Alignment Analysis}, the subspace similarity is consistently high across all layers, with an average similarity of 0.908. Even in the layers with the lowest overlap, the similarity score remains above 0.88. This strong alignment confirms that the principal directions extracted by $W_0 C$ are nearly identical to those of the theoretical solution.

Furthermore, given the phenomenon observed in Section \ref{sec:appendix_stability} (Figure \ref{fig:spectrum_analysis}), using $C$ (variance-weighted) rather than $C^{1/2}$ (standard deviation-weighted) proves more effective at suppressing the influence of noise directions and is therefore retained.
\section{Experiment Setups}
\label{sec:hyperparameters}
To ensure reproducibility, we fixed the random seed to 42 for all experiments. Due to computational constraints, all reported results are derived from a single run.
\subsection{GLUE benchmark}
\begin{table}[H]
\small
\centering
\begin{tabular}{@{}ccllllll@{}}
\toprule
Hyperparameters & \multicolumn{7}{c}{GLUE}               \\ \midrule
Rank r          & \multicolumn{7}{c}{32}                 \\
$\alpha$               & \multicolumn{7}{c}{32}                 \\
Dropout         & \multicolumn{7}{c}{0}                  \\
Optimizer       & \multicolumn{7}{c}{AdamW}              \\
LR              & \multicolumn{7}{c}{3e-4}               \\
LR Scheduler    & \multicolumn{7}{c}{Cosine}             \\
Batch size      & \multicolumn{7}{c}{32}                 \\
Warmup ratio    & \multicolumn{7}{c}{0.03}               \\
Epochs          & \multicolumn{7}{c}{3}                  \\
Where           & \multicolumn{7}{c}{q, k, v, o, wi, wo} \\ 
Sample size(TLoRA)          & \multicolumn{7}{c}{32}                  \\
\bottomrule
\end{tabular}
\caption{Hyperparameter configurations of TLoRA and other PEFT methods for T5-Base on the GLUE Benchmark tasks.}
\label{Hyperparameters GLUE}

\end{table}
\subsection{Commonsense reasoning}
\begin{table}[H]
\small
\centering
\begin{tabular}{@{}ccllllll@{}}
\toprule
Hyperparameters & \multicolumn{7}{c}{Commonsense Reasoning}                                              \\ \midrule
Rank r          & \multicolumn{7}{c}{32}                                                                 \\
$\alpha$               & \multicolumn{7}{c}{32}                                                               \\
Dropout         & \multicolumn{7}{c}{0}                                                                  \\
Optimizer       & \multicolumn{7}{c}{AdamW}                                                              \\
LR              & \multicolumn{7}{c}{1e-4}                                                               \\
LR Scheduler    & \multicolumn{7}{c}{linear}                                                             \\
Batch size      & \multicolumn{7}{c}{32}                                                                 \\
Warmup ratio    & \multicolumn{7}{c}{0.03}                                                               \\
Epochs          & \multicolumn{7}{c}{1}                                                                  \\
Where           & \multicolumn{7}{c}{q, k, v, up, down, o, gate} \\ 
Sample size(TLoRA)          & \multicolumn{7}{c}{32}       \\
\bottomrule
\end{tabular}
\caption{Hyperparameter configurations of TLoRA and other PEFT methods for Llama2-7B on the Commonsense reasoning.}
\label{Hyperparameters Commonsense}
\end{table}
\subsection{Math Code Chat}
\begin{table}[H]
\small
\centering
\begin{tabular}{@{}ccllllll@{}}
\toprule
Hyperparameters & \multicolumn{7}{c}{Math Code Chat}                                                     \\ \midrule
Rank r          & \multicolumn{7}{c}{128}                                                                \\
$\alpha$         & \multicolumn{7}{c}{128}                                                          \\
Dropout         & \multicolumn{7}{c}{0}                                                                  \\
Optimizer       & \multicolumn{7}{c}{AdamW}                                                              \\
LR              & \multicolumn{7}{c}{2e-5}                                                               \\
LR Scheduler    & \multicolumn{7}{c}{cosine}                                                             \\
Batch size      & \multicolumn{7}{c}{128}                                                                \\
Warmup ratio    & \multicolumn{7}{c}{0.03}                                                               \\
Epochs          & \multicolumn{7}{c}{1}                                                                  \\
Where           & \multicolumn{7}{c}{q, k, v, up, down, o, gate} \\ 
Sample size(TLoRA)          & \multicolumn{7}{c}{32}       \\
\bottomrule
\end{tabular}
\caption{Hyperparameter configurations of TLoRA and other PEFT methods for Llama2-7B on the Math reasoning, Code generation, and chat generation.}
\label{Hyperparameters Math}

\end{table}
\section{Analysis TLoRA}
In this section, we conducted comprehensive analyses to validate the effectiveness of TLoRA. Unless otherwise specified, all analyses in this section are evaluated based on the fine-tuning results of the Llama-2-7B model on mathematical reasoning tasks.

\subsection{Comparison with More Models}
\label{app: more_model}
\noindent\textbf{Models and Datasets.} We further assessed TLoRA's robustness and scalability through experiments on three additional models: a larger model (LLaMA2-13B \cite{touvron2023llama}), a more advanced model(LLaMA3-8B), and a model from a different architecture (Mistral-7B \cite{jiang2023mistral}). For these supplementary evaluations, we fine-tune the model on the same 100K subset of MetaMathQA \cite{yu2023metamath} from our main experiments, focusing on the mathematical reasoning domain as an efficient yet challenging performance benchmark. Performance is evaluated on the GSM8K \cite{cobbe2021training} and MATH \cite{hendrycks2021measuring}. Hyperparameters are kept consistent with the LLaMA2-7B setup. Detailed hyperparameter configurations for experiments are provided in Table~\ref{Hyperparameters Math}. 

\begin{table}[t]
\centering
\
\begin{tabular}{@{}cccc@{}}
\toprule
\setlength{\tabcolsep}{3pt}
\textbf{Model} & \textbf{method}  & \textbf{GSM8K}  & \textbf{MATH}   \\ \midrule
\multirow{4}{*}{\centering Mistral-7B} & LoRA   & 70.81 & 19.40      \\
                                      & DoRA   & 70.58 & 18.68 \\
                                      & PISSA  & 73.16 & 19.64 \\
                                       & TLoRA  & \textbf{73.38}  &  \textbf{20.32}     \\ \midrule
\multirow{4}{*}{\centering LLaMA2-13B} & LoRA   & 56.10 & 9.84      \\
                                      & DoRA   & 57.16 & 9.78      \\
                                      & PISSA  & 62.92 & 12.24     \\
                                      & TLoRA  & \textbf{64.51}  & \textbf{12.74}       \\ \bottomrule
\multirow{4}{*}{\centering LLaMA3-8B} & LoRA   & 71.94 & 9.84      \\
                                      & DoRA   & 72.10 & 22.22      \\
                                      & PISSA  & 76.26 & 24.06     \\
                                      & TLoRA  & \textbf{77.86} & \textbf{24.76}      \\ \bottomrule
\end{tabular}
\caption{Results of fine-tuning Mistral-7B, LlaMA2-13B, and LlaMA3-8B using TLoRA on a 100K subset of MetaMathQA. Bold numbers indicate the best performance achieved on this subtask.}
\label{tab:supplementary_results}
\end{table}

\noindent\textbf{Results.} Table~\ref{tab:supplementary_results} presents the comparative results across Mistral-7B, LLaMA2-13B, and LLaMA3-8B. TLoRA consistently outperforms all baseline methods—LoRA, DoRA, and PISSA—across both GSM8K and MATH benchmarks for all three models. Notably, TLoRA demonstrates exceptional scalability on the larger LLaMA2-13B model, achieving 64.51\% on GSM8K and 12.74\% on MATH, surpassing the strong baseline PISSA by clear margins. Furthermore, on the more advanced LLaMA3-8B and architecturally distinct Mistral-7B, TLoRA maintains its leadership position, achieving the highest accuracy in every setting. These results strongly validate that TLoRA's effectiveness is robust across different model scales and architectural designs, consistently delivering superior fine-tuning performance compared to existing  PEFT methods.

\subsection{Comparison with Different initialization settings.}
To validate the effectiveness of TLORA initialization, we conducted comparative ablation experiments across three distinct initialization settings. Specifically, while maintaining consistent subsequent training settings (i.e., freezing matrix $A$ and training only matrix $B$), we compared the following three strategies for constructing feature extractor $A$:
\begin{itemize}[leftmargin=*]
    \item \textbf{Random-init.} Using the standard LoRA configuration with random initialization of $A$.
    \item \textbf{Weight-only, W-SVD.} Perform SVD decomposition on the pre-trained weights $W_0$ and initialize $A$ using the first $r$ principal singular vectors.
    \item \textbf{TLoRA-init, WC-SVD.} Follow the TLoRA-init initialization matrix $A$.
\end{itemize}

\textbf{Analysis.} As shown in Table \ref{tab:init_setting}, random-init performed the worst,  which validates our theoretical motivation. Since the weight updates are strictly confined within the subspace spanned by the row vectors of $A$, a random initialization makes it hard to capture task-relevant feature directions, severely bottlenecking the model's adaptive capacity even if $B$ is optimized. Second, while W-SVD improves performance by over 10\% compared to the random baseline, it remains significantly inferior to the full TLoRA initialization. This gap suggests that weight magnitude in isolation is a suboptimal proxy for importance. Finally, WC-SVD (TLoRA) achieves the best results, outperforming W-SVD by a substantial margin of 8.34\% on GSM8K and 1.90\% on MATH. This empirical success confirms that the weighted product $W_0C$ more accurately captures the principal components of the task-specific update subspace, allowing the model to achieve superior adaptation performance with a frozen feature extractor.
\begin{table}[ht]
    \centering
    \begin{tabular}{ccccccc}
   \toprule
    \multicolumn{1}{c}{\textbf{Init}} & \textbf{GSM8K}  & \textbf{Math}   \\ \midrule
    Random-init                                & 33.20         & 4.86        \\
    W-SVD                                & 43.44          & 5.84         \\
    WC-SVD                                & 51.78           & 7.74         \\\bottomrule
\end{tabular}
\caption{Results of fine-tuning llama2-7B using different init setting.}
\label{tab:init_setting}
\end{table}

\subsection{Sensitivity Analysis of Sample Size}
Our task-aware initialization is a data-driven initialization method. A crucial issue is the sensitivity of our method to the number of samples used during initialization. To evaluate the robustness of TLoRA, we conduct a sensitivity analysis by varying the amount of data used for initialization and observing the impact on the performance of the final model.

\noindent\textbf{Analysis.} The results in the table \ref{tab:sample_analysis} indicate that the performance of TLoRA is very robust to the number of samples used for calibration. In GSM8K and MATH benchmark tests, whether using small batches (such as 16 samples) or large batches (such as 512 samples), the final accuracy remains within a very narrow and high-performance range. This discovery strongly indicates that our task-aware initialization can effectively estimate high-quality adaptive subspaces from a small number of representative samples without requiring a large amount of calibration data. This result highlights the stability and practicality of TLoRA.

\begin{table}[h]
\centering
\begin{tabular}{@{}ccc@{}}
\toprule
method                  & GSM8K & MATH \\ \midrule
TLoRA(with 16 samples)  & 56.17 & 8.70 \\
TLoRA(with 32 samples)  & 56.34 & 9.08 \\
TLoRA(with 64 samples)  & 56.64 & 9.00 \\
TLoRA(with 128 samples) & 56.56 & 8.98 \\
TLoRA(with 256 samples) & 56.94 & 8.88 \\
TLoRA(with 512 samples) & 55.88 & 9.12 \\ \bottomrule
\end{tabular}
\caption{Results of fine-tuning LlaMA2-7B using different sample size initialization TLoRA on a 100K subset of MetaMathQA.}

\label{tab:sample_analysis}
\end{table}

\subsection{Robustness of TLoRA towards different ranks}
This section investigates the effect of varying rank configurations on the performance of TLoRA and LoRA. We evaluate the performance of the fine-tuned LLaMA2-7B model on math reasoning and code generation tasks. Table \ref{tab:diff_rank} illustrates the performance of both methods across different rank settings. Notably, TLoRA consistently outperforms LoRA in all configurations, demonstrating its effectiveness in improving fine-tuning performance and its robust generalization capabilities.

\begin{table}[ht]
\centering
\small
\setlength{\tabcolsep}{2pt}
\begin{tabular}{lcccccc}
\toprule
\textbf{Method} & \textbf{Rank} & \textbf{GSM8K} & \textbf{MATH} & \textbf{MBPP} & \textbf{HumanEval} \\
\midrule
\multirow{4}{*}{\centering LoRA}    
    & 16  & 32.90 & 4.22 & 32.80 & 15.90  \\ 
    & 32  & 37.07  & 4.54 & 34.10 & 16.50  \\
    & 64  & 40.48  & 5.42 & 36.50 & 17.70   \\ 
    & 128 & 44.80  & 6.18 & 35.70 & 20.70  \\ 
\midrule
\multirow{4}{*}{\centering TLoRA}    
    & 16  & 44.90 & 6.20 & 34.30 & 18.90   \\ 
    & 32  & 47.80 & 6.64 & 37.30 & 19.50  \\
    & 64  & 52.87 & 7.92 & 39.00 & 22.60  \\ 
    & 128 & 56.35 & 9.08 & 39.20 & 23.50  \\ 
\bottomrule
\end{tabular}
\caption{Comparison of LoRA and TLoRA with varying ranks for LLaMA2-7B on different tasks.}
\label{tab:diff_rank}
\end{table}
\section{Differentiated Module Importance Across Tasks} 
\begin{figure}[H]
    \centering
    \includegraphics[width = 0.99\linewidth]{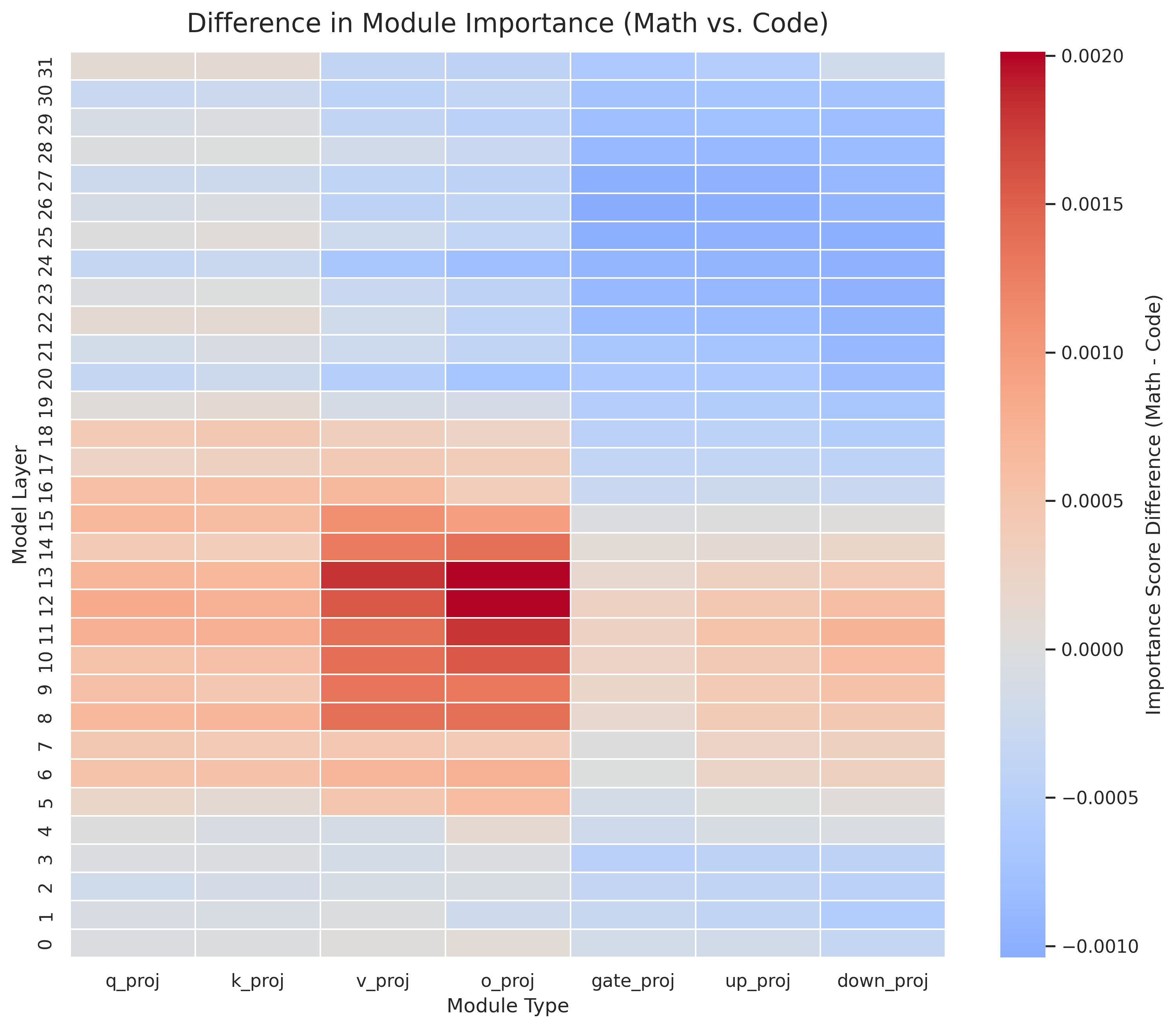}
    \caption{The data on the graph shows the difference in importance scores between mathematical tasks and code tasks.}
    \label{fig:score}
\end{figure}

To further probe the task-specific nature of module importance, we analyse the difference in scores between the mathematical reasoning and code generation domains. We subtract the importance score of each module on the code task from its importance score on the math task. The resulting importance difference, visualised in a heatmap in Figure \ref{fig:score}, reveals a highly localised pattern rather than a global shift. We observe a distinct difference in the middle layers of the network (approximately layers 8-15), where the $v_{proj}$ and $o_{proj}$ modules exhibit significantly higher importance for math reasoning than for code generation. In contrast, the vast majority of other modules across all layers show negligible differences, suggesting their roles are largely task-agnostic.

This finding—that only a sparse, specific subset of modules is highly task-sensitive—powerfully underscores the necessity of an adaptive allocation strategy like TLoRA. It demonstrates that superior performance hinges on the ability to identify and concentrate the parameter budget onto these few critical, task-specific modules, rather than uniformly distributing it across the many general-purpose ones.

\section{Baseline}
\noindent\textbf{Baselines.} We compare TLoRA with several baselines in Experiments:
\begin{itemize}[leftmargin=*]
    \item \textbf{FULL} Fine-tuning the model with all parameters.
    \item \textbf{LoRA}\cite{hu2022lora} approximates weight $\Delta W$ updates through the product of two trainable low-rank matrices $A$ and $B$.
    \item \textbf{AdaLoRA}\cite{zhang2023adalora} enhances performance by dynamically allocating the parameter budget across layers based on their importance.
    \item\textbf{LoRA+}\cite{hayou2024lora+} improves upon LoRA by setting a higher learning rate for the adapter matrix $B$ than for matrix $A$.
    \item \textbf{DoRA}\cite{liu2024dora} decompose the pretrained weight matrix $W$ into magnitude and direction component.
    \item \textbf{PiSSA}\cite{meng2024pissa} initializes adapter matrices $A$ and $B$ using the principal singular components of the original weight matrix $W$. 
    \item \textbf{OLoRA}\cite{buyukakyuz2024olora} initializes adapter matrices $A$ and $B$ using the orthogonal bases of the original weight matrix $W$. 
    \item \textbf{LoRA-GA}\cite{wang2024lora} aligns the low-rank adaptation with the gradient direction of the pre-trained weights by using a gradient-based approximation to initialize matrix $A$ and $B$.
    \item \textbf{CorDA}\cite{yang2024corda} constructs a task-relevant covariance matrix to identify important directions in the weight space, initializing the adapter based on the singular vectors of these covariance-weighted features.
\end{itemize}

\end{document}